\documentclass{article} 
\usepackage{iclr2016_workshop,times}
\usepackage{hyperref}
\usepackage{url}
\usepackage{amsmath}
\usepackage{amssymb}
\usepackage[pdftex]{graphicx}
\usepackage[font=small,labelfont=bf]{caption}

\title{Learning Genomic Representations to Predict Clinical Outcomes in Cancer}
\author{Safoora Yousefi$^{\dagger}$ \qquad Congzheng Song$^{\dagger}$ \qquad
Nelson Nauata$^{\ddagger}$ \qquad
Lee Cooper$^{\ddagger,\divideontimes}$
\\
$^{\dagger}$Department of Mathematics \& Computer Science
Emory University\\
$^{\ddagger}$Department of Biomedical Informatics 
Emory University School of Medicine\\
$^{\divideontimes}$Department of Biomedical Engineering
Georgia Institute of Technology\\
\texttt{\{safoora.yousefi, congzheng.song, nnauata, lee.cooper\}@emory.edu}
}
\begin{document}
\maketitle

\vspace{-3mm}
\begin{abstract}
Genomics are rapidly transforming medical practice and basic biomedical research, providing insights into disease mechanisms and improving therapeutic strategies, particularly in cancer. The ability to predict the future course of a patient's disease from high-dimensional genomic profiling will be essential in realizing the promise of genomic medicine, but presents significant challenges for state-of-the-art survival analysis methods. In this abstract we present an investigation in learning genomic representations with neural networks to predict patient survival in cancer. We demonstrate the advantages of this approach over existing survival analysis methods using brain tumor data.
\end{abstract}

\vspace{-3mm}
\section{Introduction}
Genomics provide a window into the complex molecular workings of disease. In the treatment of cancer, genomic analysis of a tissue biopsy can reveal specific molecular vulnerabilities that can be matched to targeted therapies, or to prognosticate the future behavior of a patient's disease and expected survival in order to better inform clinical interventions including surgery and radiation therapy. Although genomic analysis generates rich high-dimensional signals that contain hundreds to hundreds-of-thousands of variables, typically only several variables are used for prognostication for any given cancer type. Typically, these variables are used to assign patients into discrete disease classes or "subtypes" that associate with response to specific therapies, or with varying degrees of disease aggressiveness. Learning the underlying latent prognostic variables from high-dimensional genomic profiles can extract additional prognostic value from unused variables, and is critical in realizing the promise of genomic medicine. This problem presents significant challenges, ranging from the familiar "large \emph{p} small \emph{N}", to how to adapt developments in the machine learning domain to the analysis of time-to-event survival data.

In this abstract we present an investigation in building survival prediction neural networks to learn representations from genomic data for survival prediction. We use backpropagation to train neural networks to maximize the Cox proportional hazards likelihood of time-to-event data, and apply these predictive models to molecular profiles of brain tumor patients from The Cancer Genome Atlas where survival ranges from 6 months to 10+ years. We compare our methods to state-of-the-art survival analysis algorithms based on elastic-net (linear combination of L1 and L2) regularization of Cox hazard models and random forest based methods, and demonstrate improvements in survival prediction accuracy for neural network approaches. 

\section{Background and Related Work}
\subsection{Hazard models and likelihood functions}\vspace{-3mm}
Survival analysis involves predicting the time to some event of interest, which in cancer is often death or progression of disease. It differs from ordinary regression due to \emph{incomplete followup}, where a death or relapse event is not observed at or before the final encounter with the patient. These censored observations provide critical information, and often represent an important population of long-term survivors or treatment responders that are very important to incorporate into the model.\\ 
The most commonly used regression approach to survival analysis is the Cox proportional hazards model proposed by \citet{david1972regression}. At time $t$, the hazard for a sample with covariates $x$ is given by the following hazard function:
\begin{equation}
    \lambda(t|x) = \lambda_0(t)  e^{\beta x},
\end{equation}
where $\lambda_0(t)$ is baseline hazard. The hazard function is an exponential linear function of the covariates $x$ and model coefficients $\beta$, with the effect of any covariate $x^{(i)}$ assumed to be the same over time.
Since the evaluation criteria of the models in this paper is based on ranking predicted survival times, the baseline hazard is left unspecified and we maximize the partial likelihood function during training:
\begin{equation}
    l(\beta, X) = - \sum_{i\in U}{\Big(X_i\beta - \log\sum_{j\subset{R_i}}{e^{X_j\beta}}\Big)}
    \label{eq:costfunc}
\end{equation}
where $U$ is the set of all uncensored patients, and $R_i$ is the set of patients whose time of death or last follow-up is later than time of death of $i$.

We measured model performance using \emph{concordance index} (CI) that captures the rank correlation of predicted and actual survival. Denoting the $i$th patient with $X_i$ and the set of all patients with $X$, where $t_i$ represents either the time of death or the time of last follow-up of the $i$th patient, CI was calculated in the following way:
\vspace{-2mm}
\begin{equation}\vspace{-2mm}
    \textit{CI}(\beta, X) =  \sum_P{\tfrac{I(i, j)}{|P|}}
\end{equation}
\begin{equation}
    I(i, j) = 
\begin{cases}
    1,& \text{if } Risk_j > Risk_i\ \text{and}\ t_j > t_i\\
    0,              & \text{otherwise}
\end{cases}
\end{equation}

Where $P$ is the set of orderable pairs. A pair of samples ($X_i$, $X_j$) is orderable if either the event is observed for both $X_i$ and $X_j$, or $X_j$ is censored and $t_j > t_i$.
Intuitively, CI measures the pairwise agreement of the prognostic scores $Risk_i$, $Risk_j$ predicted by the model and the actual time of death for all orderable pairs. Attempts have been made to propose differentiable versions of CI based on sigmoid and exponential functions and optimize it directly. But recent studies show that optimizing the cox partial likelihood is equivalent to optimizing CI \citep{steck2008ranking}.

\subsection{Related Works}\vspace{-3mm}
Regularization techniques have been proposed for feature selection in survival analysis of high dimensional data \citep{zou2005regularization}. Efforts have been made to introduce successful machine learning algorithms such as random forests to survival analysis \citep{ishwaran2008random}. Deep learning techniques have been employed for cancer diagnosis using genomic data and medical images, such as \citet{fakoor2013using} and \citet{estevadeep} To the best of our knowledge, representation learning techniques have not been applied to survival prediction from genomic data, and the previous work investigating neural networks for survival analysis dealt with low dimensional data and different cost functions \citep{lisboa2003bayesian}.

\section{Survival Prediction Neural Network}
\subsection{Pretraining and fine-tuning}
\begin{figure}[h]
\centering
\begin{tabular}{c c}
\includegraphics[width=0.42\linewidth]{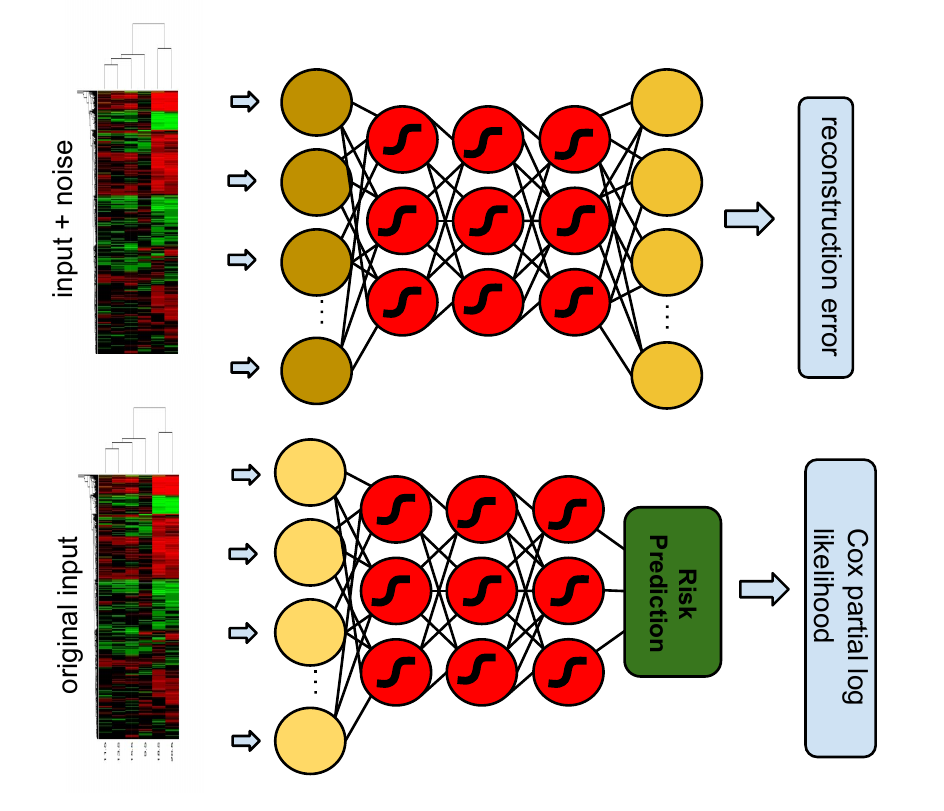} &
\includegraphics[width=0.49\linewidth]{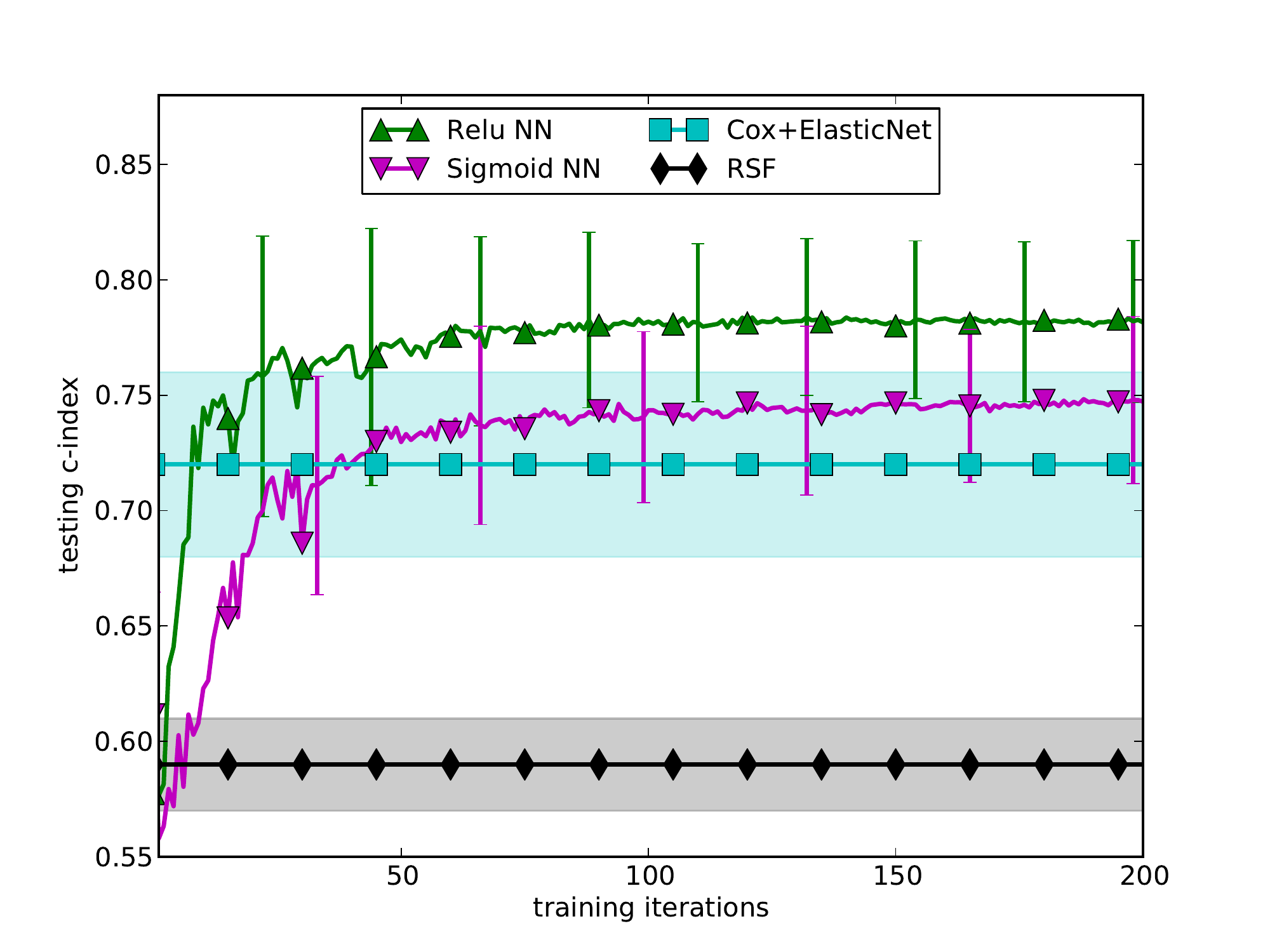} \\
a) proposed network & b) c-index results
\end{tabular}
\caption{a) Survival network model. b) Comparison of the proposed survival prediction neural network to two competing methods: Elastic-net (L1 and L2 combined) regularized Cox regression and Random Survival Forest(RSF). Average testing CI trends are shown for neural networks with two different activation functions: sigmoid and rectified linear units. The error bars and shaded areas indicate standard deviation of CI over 10 cross validation sets (See section \ref{evaluation} for details.)}
\label{fig:coxnn-result}
\end{figure}

\vspace{-3mm}
In this work we trained an autoencoder to represent genomic data and fine tune this representation using partial log Cox likelihood. In training, we employ stacked denoising autoencoders proposed in \citet{vincent2008extracting}. We train the auto-encoders using 183-dimensional genomic features, then we add a risk prediction output layer as shown in Figure \ref{fig:coxnn-result}-a. 


We use survival times and censoring status to calculate the Cox partial log likelihood given by equation \ref{eq:costfunc} and differentiate it with respect to $X$:
\begin{equation}
	\frac{\partial{l(\beta, X)}}{\partial{X_i}} = c_i\beta\ -\sum_{j\in U:i\in R_j}{\frac{\beta e^{X_i\beta}}{\sum_{k\in R_j}{e^{X_k\beta}}}}
\label{eq:derivative}
\end{equation}

where $c_i$ is $1$ if sample $i$ is not censored, and is $0$ otherwise, and $\beta$ denotes the parameters of the risk prediction layer. This derivative is then back-propagated through the network to fine tune the learned representation specifically for the task of survival analysis.

\subsection{Model Selection}\vspace{-3mm}
The training of a neural network involves many hyperparameters: type of nonlinearities used, number of layers, number of hidden units in each layer, learning rates for pretraining and fine tuning and regularization parameters. Since this is the first work where deep neural networks are used to address survival analysis, we could not look at existing literature for a conventional choice of hyperparameters. Unlike areas such as image classification, no rule of thumb has been developed for setting hyper-parameters in survival analysis. Therefore we employed bayesian optimization \citep{martinez2014bayesopt} with Gaussian prior to decrease the number of objective function evaluations needed to reach a decent choice of hyperparameters. 
More shallow networks demonstrate superior performance over deeper architectures in our experiments. This could be justified considering the small number of training samples (628) and the scarcity of labels within the available samples. Our average choice of configuration is 2 fully connected layers of 250 hidden units each. On average, we use a learning rate .001 for pre-training and .0009 for fine-tuning.

\subsection{Evaluation}\vspace{-3mm}
\label{evaluation}
Due to the small size of available training data, performance of the model might considerably depend on the partitioning of the data into testing and training. To mitigate this, we randomly sampled from the data set $10$ times without replacement to have 10 permutations of the same data set. Then in each of the $10$ sets, we used the first $\%70$ of the data for training, half of the remaining $\%30$ of data for model selection and the other half for model assessment. We performed training, model selection and testing on these 10 permutation datasets separately. The reported CI in Figure\ref{fig:coxnn-result}-b is averaged over these 10 experiments to represent the generalization error of the model.
The exact same setting was used for hyper-parameter tuning and assessment for competing methods. We picked the learning rate and elastic-net mixture coefficient for regularized Cox regression (\citet{hastie2014glmnet}) based on performance on the same validation sets we used for the neural networks. We tuned number of trees, leaf size, and number of split points for random survival forest in the same fashion.
Our experiments reveal that Random Survival Forests do not adapt well to high dimensionality and are markedly outperformed by survival neural networks (See Figure \ref{fig:coxnn-result}-b). Neural networks also achieve $\%5$ absolute improvement over regularized Cox regression with ReLU activation and $\%3$ with sigmoid activation.

\section*{Acknowledgments}\vspace{-3mm}
This work was supported by US Public Health Service National Institutes of Health (NIH) grants K22LM011576-03 and U24CA194362-01.
\bibliography{iclr2016_workshop}
\bibliographystyle{iclr2016_workshop}

\end{document}